\documentclass{article}
\usepackage{ijcai26}

\usepackage{times}
\usepackage{soul}
\usepackage{url}
\usepackage[hidelinks]{hyperref}
\usepackage[utf8]{inputenc}
\usepackage[small]{caption}
\usepackage{graphicx}
\usepackage{amsmath}
\usepackage{amsthm}
\usepackage{booktabs}
\usepackage{algorithm}
\usepackage{algorithmic}
\usepackage[switch]{lineno}
\usepackage{natbib}
\usepackage{tcolorbox}

\usepackage{siunitx}
\usepackage{booktabs}
\usepackage{multirow}
\usepackage{float}
\usepackage{pifont}

\title{Discourse-Aware Dual-Track Streaming Response for \\ Low-Latency Spoken Dialogue Systems}

\author{
Siyuan Liu$^{1, 5}$ \and
Jiahui Xu$^{2}$ \and
Feng Jiang$^{3}$ \and
Kuang Wang$^{4}$ \and \\
Zefeng Zhao$^{4}$ \and 
Chu-Ren Huang$^{2}$ \and
Jinghang Gu$^{2}$ \and
Changqing Yin$^{1}$ \and
Haizhou Li$^{5}$ \\
\affiliations
$^{1}$School of Computer Science and Technology, Tongji University \\
$^{2}$Department of Language Science and Technology, Hong Kong Polytechnic University \\
$^{3}$Artificial Intelligence Research Institute, Shenzhen University of Advanced Technology \\
$^{4}$School of Data Science, Chinese University of Hong Kong, Shenzhen \\
$^{5}$School of Artificial Intelligence, Chinese University of Hong Kong, Shenzhen \\
\emails
2433259@tongji.edu.cn,
jiangfeng@suat-sz.edu.cn \\
}

\begin{document}

\maketitle

\begin{abstract}
Achieving human-like responsiveness is a critical yet challenging goal for cascaded spoken dialogue systems. Conventional ASR–LLM–TTS pipelines follow a strictly sequential paradigm, requiring complete transcription and full reasoning before speech synthesis can begin, which results in high response latency. We propose the Discourse-Aware Dual-Track Streaming Response (DDTSR) framework, a low-latency architecture that enables listen-while-thinking and speak-while-thinking. DDTSR is built upon three key mechanisms: (1) connective-guided small–large model synergy, where an auxiliary small model generates minimal-committal discourse connectives while a large model performs knowledge-intensive reasoning in parallel; (2) streaming-based cross-modal collaboration, which dynamically overlaps ASR, LLM inference, and TTS to advance the earliest speakable moment; and (3) curriculum-learning-based discourse continuity enhancement, which maintains coherence and logical consistency between early responses and subsequent reasoning outputs. Experiments on two spoken dialogue benchmarks demonstrate that DDTSR reduces response latency by 19\%–51\% while preserving discourse quality. Further analysis shows that DDTSR functions as a plug-and-play module compatible with diverse LLM backbones, and remains robust across varying utterance lengths, indicating strong practicality and scalability for real-time spoken interaction.
\end{abstract}

\section{Introduction}

Spoken dialogue involves real-time interaction conducted through speech, where interlocutors continuously exchange information and coordinate intent~\citep{lopez2014review, park2024let}. Natural turn-taking typically requires response onsets within a few hundred milliseconds across languages~\citep{stivers2009turntaking, heldner2010pauses, levinson2015timing}. 
Such stringent latency requirements are fundamental to human communication and critically shape user experience in applications ranging from conversational assistants to accessibility technologies.

\begin{figure}[t]
\centering
\includegraphics[width=\linewidth]{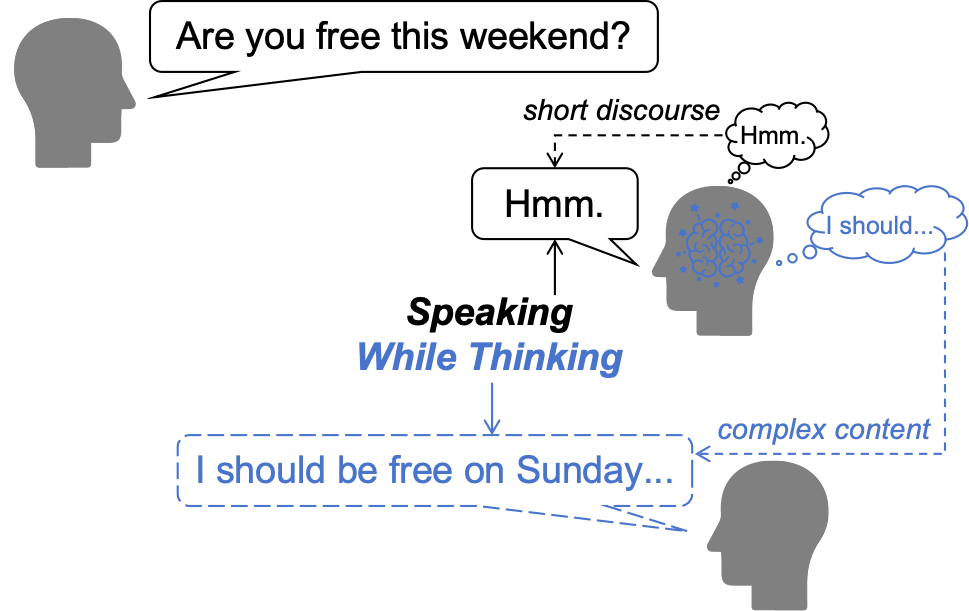}
\caption{Necessity of low-latency spoken interaction. Humans naturally exhibit speaking while thinking behavior, whereas conventional systems wait for full response completion before speaking.}
\label{fig:speaking_while_thinking}
\end{figure}

Recent end-to-end Speech models~\citep{ji2024wavchat, fang2024llamaomni, xu2025qwen3omni, defossez2024moshi} have shown strong potential by directly mapping speech to speech, simplifying system pipelines and reducing intermediate processing. 
However, such monolithic architectures remain computationally expensive, data-intensive, and tightly coupled, requiring specialized training pipelines and limiting modular deployment or component-level replacement. 
Consequently, many practical and reproducible spoken dialogue systems, such as FireRedChat~\citep{chen2025fireredchat} and X-Talk~\citep{liu2025x}, continue to rely on the cascaded paradigm for its modularity, controllability, and ease of integration.

Despite these advantages, achieving human-like responsiveness within cascaded systems remains a central challenge. 
In conventional pipelines, recognition, reasoning, and synthesis are executed strictly in sequence, forcing speech output to wait for complete transcription and full semantic generation. 
This sequential design pushes response onset far beyond the sub-second regime required for natural interaction, leading to high perceived latency and disrupted conversational flow. 
This tension exposes a fundamental dilemma:  
\textbf{How can we preserve the modular benefits of cascaded architectures while substantially reducing perceived response latency?}

Psycholinguistic studies suggest that humans naturally speak while thinking~\citep{levelt1989speaking, clark2002uhum, bogels2015neural}. 
In everyday conversation, speakers frequently initiate responses with short discourse cues such as “well,” or “I see” to maintain conversational flow while higher-level planning continues in parallel. 
These linguistic devices, often referred to as discourse connectives~\citep{dorgeloh2022discourse} or discourse markers~\citep{fraser2009account}, function as low-risk interactional buffers, as they typically do not convey truth-conditional content and therefore avoid knowledge-level conflicts with the subsequent full response generated by a large model, allowing response onset to occur before full content formulation.

Inspired by this principle, we propose the \textbf{Discourse-Aware Dual-Track Streaming Response (DDTSR)} framework, a novel low-latency architecture for cascaded spoken dialogue systems~\footnote{The code is released at \url{https://github.com/hlt-cuhksz/DDTSR}.} 
Rather than redefining dialogue protocols or relying on fully end-to-end models, DDTSR targets the temporal structure of interaction within cascaded pipelines. 
Its core idea is to explicitly decouple early, minimal-committal speech generation from knowledge-intensive reasoning, enabling the system to listen while thinking and speak while thinking.

DDTSR is built upon three tightly integrated mechanisms:

\textbf{(1) Connective-Guided Small–Large Model Synergy.}  
We decompose response generation into two complementary roles to minimize perceived thinking time: a lightweight model generates discourse connectives that are safe for immediate emission, while a large model performs deep reasoning in parallel.

\textbf{(2) Streaming-Based Cross-Modal Collaboration.}  
Instead of waiting for finalized ASR output, DDTSR incrementally consumes partial transcripts and allows ASR, LLM, and TTS to operate in an interleaved manner. 
This cross-modal temporal reordering effectively masks reasoning latency and advances the earliest speakable moment.

\textbf{(3) Curriculum-Learning-Based Discourse Continuity Enhancement.}  
To ensure that early connectives remain discourse compatible with subsequent reasoning results, we introduce coherence and consistency-aware training objectives and a confidence-driven emission policy, enabling early speech that is both responsive and discourse-continuous.

Experimental results on two widely used spoken dialog benchmarks show that, compared to conventional single-stream cascaded systems, DDTSR achieves a 19–51\% reduction in response latency and consistently reaches sub-second response onset. Our main contributions are as follows:

\begin{itemize}
\item We formulate low-latency spoken dialogue as a temporal reorganization problem and identify the response-onset dilemma in cascaded ASR--LLM--TTS systems.

\item We propose DDTSR, a discourse-aware dual-track streaming framework that decouples early connective generation from deep reasoning via small–large model synergy and cross-modal streaming collaboration.

\item Experiments demonstrate that DDTSR as a plug-and-play module reduces perceived and end-to-end latency by 19\%–51\% while preserving response quality with strong practicality and scalability.
\end{itemize}

\section{Related Work}

\subsection{Cascaded Spoken Dialogue System}

Most practical spoken dialogue systems adopt a cascaded ASR–LLM–TTS architecture, in which speech recognition, language understanding, and speech synthesis are implemented as independent modules.
This design provides strong modularity, ease of integration, and component-level replaceability, and therefore underpins a wide range of deployed systems~\citep{radford2023whisper, kim2021vits, arora2025espnet, kennington2025incremental}.
Despite these advantages, cascaded pipelines exhibit a fundamental limitation for natural spoken interaction when compared with end-to-end speech models~\citep{ji2024wavchat, fang2024llamaomni, xu2025qwen3omni, defossez2024moshi}: their strictly sequential execution accumulates latency across recognition, reasoning, and synthesis stages.
As a result, response onset is often delayed far beyond the sub-second timescale required for human-like turn-taking~\citep{stivers2009turntaking, heldner2010pauses}.

\subsection{Latency Reduction in Cascaded Systems}

\paragraph{Faster computation is not faster interaction.}
A large body of work aims to reduce generation latency by accelerating model computation.
Speculative decoding and draft–verify strategies allow lightweight models to propose candidate tokens that are selectively verified by stronger models, improving decoding throughput while preserving output quality~\citep{leviathan2023speculative, chen2023speculativesampling, liu2024onlinespecdec}. 
Multi-token prediction frameworks further amortize decoding cost via auxiliary prediction heads~\citep{cai2024medusa}. 
While these approaches effectively reduce per-token inference cost, they primarily optimize computational latency rather than interaction latency. 
Even with faster decoding, response onset in cascaded systems remains fundamentally constrained by their sequential structure, where speech synthesis can begin only after LLM reasoning has completed.

\paragraph{Early response and incremental generation.}
Another line of work seeks to reduce perceived latency by enabling earlier partial responses.
Incremental dialogue processing streams intermediate hypotheses across modules so that downstream generation can begin before upstream processing is fully complete~\citep{schlangen2009ageneral, skantze2009incremental, buss-etal-2010-collaborating, dethlefs2012optimising}. 
Related systems~\citep{baumann-schlangen-2012-inprotk, ma-etal-2020-incremental, dang2024livespeech} explore streaming ASR and early TTS triggering based on partial ASR or LLM outputs to shorten silence gaps and improve responsiveness. 
However, such methods still rely on content-level readiness: meaningful speech is typically produced only after sufficient semantic information becomes available. 
As a result, the earliest speakable moment remains bounded by the arrival of reliable propositional content, and latency reduction is limited by the need to wait for semantic completion.

\begin{figure*}[t]
    \centering
    \includegraphics[width=0.9\linewidth]{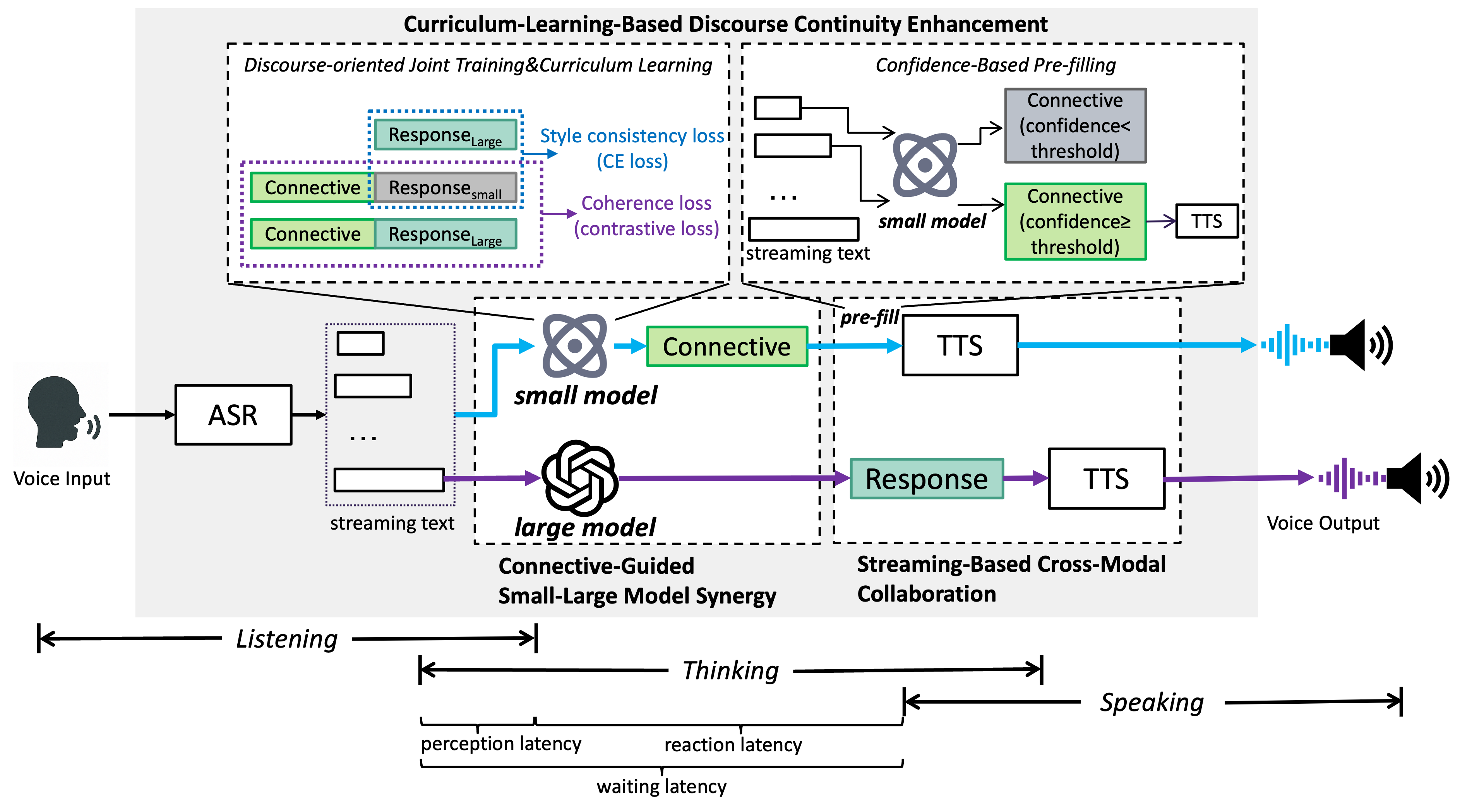}
    \caption{The Framework of Discourse-Aware Dual-Track Streaming Response.}
    \label{fig:DDTSR}
\end{figure*}

\subsection{Discourse-level Cues for Conversational Naturalness}

Linguistic and psycholinguistic studies ~\citep{schiffrin1987discourse, fraser2009account} have consistently shown that human speakers rely on non-knowledge-bearing discourse cues, such as fillers, acknowledgments, and discourse markers, to manage conversational flow while substantive reasoning is still in progress.  
These cues enable speakers to hold the conversational floor, signal engagement, and initiate responses before fully formulating their intended content, forming the cognitive basis of the natural speaking-while-thinking behavior observed in human spoken interaction.

Prior studies~\citep{liu2016recognizing, pan2018discourse, nie2019dissent} have leveraged discourse markers to analyze turn-taking behavior, predict discourse relations, and improve dialogue coherence modeling.  
These works demonstrate that such cues provide valuable information about interactional structure and conversational intent.  
However, they primarily focus on offline analysis or text-based dialogue generation, rather than exploiting discourse cues as part of the real-time system architecture. As a result, the potential of discourse connectives as operational mechanisms for low-latency spoken dialogue systems remains largely unexplored.

\section{Discourse-Aware Dual-Track Streaming Response Framework (DDTSR)}

Spoken dialogue systems inherently face a latency–quality dilemma: users expect to hear a response almost immediately after speaking, yet producing a correct and coherent answer typically requires substantial reasoning time.

Inspired by the human thinking-while-speaking behavior as mentioned above, we propose a discourse-aware dual-track streaming architecture that enables the system to listen while thinking and thinking while speaking. The central idea is to advance the earliest time at which the system can safely start speaking, while allowing deep reasoning to continue in the background.

To realize this goal, DDTSR is designed around three integrated principles, as illustrated in Figure~\ref{fig:DDTSR}:
(1) Connective-Guided Small–Large Model Synergy, (2) Streaming-Based Cross-Modal Collaboration, and (3) Curriculum-Learning-Based Discourse Continuity Enhancement.

\subsection{Connective-Guided Small--Large Model Synergy}

A fundamental challenge in cascaded spoken dialogue systems is the long reasoning time of large language models. Simply replacing the large model with a smaller one can reduce latency, but this often leads to degraded response quality and reduced factual reliability. To address this dilemma, DDTSR introduces a connective-guided small–large model synergy mechanism that decouples response generation into two roles with distinct risk profiles.

Specifically, DDTSR decomposes response generation into two complementary components: \textbf{connective generation}: producing short, discourse-appropriate connective prefixes that are semantically safe and require little or no reasoning;
\textbf{Main response generation}: producing the knowledge-intensive content that directly addresses the user query.

These two components are assigned to models of different capacities. A lightweight model $f_s(\cdot)$ is dedicated to generating low-risk connective connectives, while a large model $f_l(\cdot)$ performs full semantic reasoning. To further reduce response latency, the two models operate concurrently upon receiving the user input $\mathbf{u}$. The final spoken output is then constructed in a temporally staged manner as shown in Eq.~\ref{eq:y}, where $\oplus$ denotes sequential concatenation.

\begin{equation}\label{eq:y}
\mathbf{y} = \mathrm{TTS}(f_s(\mathbf{u})) \oplus \mathrm{TTS}(f_l(\mathbf{u})),
\end{equation}

\subsection{Streaming-Based Cross-Modal Collaboration}

To further break through the strict constraints of the conventional ASR $\rightarrow$ LLM $\rightarrow$ TTS pipeline, DDTSR restructures the temporal workflow of cascaded systems and introduces a streaming-based cross-modal collaboration mechanism. The core idea is to allow the small model to think during ASR processing, and, in the subsequent modalities after ASR, to enable the small and large models to operate in different modalities in parallel rather than sequentially. This mechanism consists of two tightly coupled components.

\paragraph{(1) Early emission during streaming ASR.}
Instead of waiting for a finalized transcript, DDTSR operates directly on partial hypotheses produced by a streaming ASR system. 
Let $\mathbf{x}$ denote the input speech stream. We use $k$ to index these streaming updates, where each step corresponds to a new ASR decoding result, as shown in Eq.~\ref{eq:u}.

\begin{equation}\label{eq:u}
\hat{\mathbf{u}}^{(k)} = \mathrm{ASR}(\mathbf{x}_{1:k}),
\qquad k = 1,2,\dots
\end{equation}

Here, $k$ denotes the $k$-th ASR decoding step, $\mathbf{x}_{1:k}$ represents the speech observed up to time $k$, and $\hat{\mathbf{u}}^{(k)}$ is the corresponding partial transcript available at that moment.

As shown in Eq.~\ref{eq:cs}, given each evolving hypothesis $\hat{\mathbf{u}}^{(k)}$, the lightweight model $f_s$ predicts two outputs: (i) a candidate connective $c^{(k)}$, and (ii) a binary commit signal $s^{(k)}$ indicating whether this connective is safe to emit. The system commits to the earliest reliable emission point by selecting the first step at which a positive commit decision is produced, as shown in Eq.~\ref{eq:ks}.

\begin{equation}\label{eq:cs}
\left(c^{(k)},\, sig^{(k)}\right)
= f_s(\hat{\mathbf{u}}^{(k)}),
\qquad
sig^{(k)} \in \{0,1\}.
\end{equation}

\begin{equation}\label{eq:ks}
k^\star =
\min \left\{ k \mid sig^{(k)} = 1 \right\}
\end{equation}

At this moment, TTS is triggered immediately to synthesize the connective $\mathrm{TTS}(c^{k^\star})$, often before the user has finished speaking. 
This mechanism enables rapid audible feedback while higher-level reasoning continues in parallel, thereby approximating natural human conversational timing.

\paragraph{(2) Parallel small–large model execution across modalities.}
Meanwhile, while the small model prepares and emits early speech, the large model performs knowledge-intensive reasoning in parallel based on the finalized ASR result:

\begin{equation}
r_t = f_l(\hat{\mathbf{u}}^{(\infty)}),
\end{equation}

where $\hat{\mathbf{u}}^{(\infty)}$ denotes the final transcript. During the playback of $\mathrm{TTS}(c^{k^\star})$, the large model continues to generate the substantive response and synthesize its corresponding speech in the background. As soon as the small-model audio finishes, the system seamlessly continues with the large-model audio.

\subsection{Curriculum-Learning-Based Discourse Continuity Enhancement}

Beyond reducing response latency, we further aim to ensure that early-emitted discourse connectives remain aligned with the final discourse trajectory. To this end, we introduce a curriculum-learning-based module for enhancing discourse continuity that explicitly aligns early connective generation with the coherence and consistency of the final response. The module consists of the following three mechanisms.

\paragraph{(1) Discourse-oriented joint training.}  
The lightweight model is trained to generate turn-transition connectives that anticipate the discourse trajectory of the subsequent large-model response. To support this training, we construct a connective-annotated dialogue dataset in $u$[user input]--$c$[connective]--$R$[response] format using a two-stage pipeline: POS-Based Connective Extraction~\citep{sileo2019mining,prasad2017penn} and LLM-Prompted Connective Generation (See Appendix~\ref{app:connective_data_construction} for details). Then, we optimize the lightweight model on this dataset to produce connectives that anticipate the large-model response, using the following three objectives.

First, a \textbf{style consistency loss} guides the lightweight model to prefer connective $c$ whose induced response semantic style, which is aligned with the anticipated continuation $R_L = (r_1, \dots, r_{|R_L|})$ generated by the large model $f_l$, using generation as an auxiliary supervisory signal for connective selection:
\begin{equation}
    \mathcal{L}_{\text{con}} = -\sum_{t=1}^{|R_L|} \log P_{f_s}(r_t \mid u, c, r_{<t}),
\end{equation}
where $r_{<t} = (r_1, \dots, r_{t-1})$ denotes the preceding tokens. Second, since the connective–response pairs generated directly by the pretrained lightweight model $f_{s_0}$ exhibit low perplexity, indicating discourse coherence, we introduce a \textbf{coherence loss} that transfers discourse coherence at the level of connective–response pairs from $f_{s_0}$ to the trained lightweight model $f_s$. This is achieved by minimizing the gap between the perplexity of the large-model continuation $R_L$ and $c$ evaluated under $f_s$, and that of the original lightweight-model response $R_S$ and $c$ evaluated under $f_{s_0}$ conditioned on the user input $u$: 
\begin{equation}
    \mathcal{L}_{\text{coh}} = \bigl(\mathrm{PPL}_{f_s}(c, R_L \mid u) - \mathrm{PPL}_{f_{s_0}}(c, R_S \mid u)\bigr)^2,
\end{equation}
where $\mathrm{PPL}_{f_s}(\cdot)$ denotes the average token-level negative log-likelihood under model $f_s$ (see Appendix~\ref{app:ppl} for details). Finally, a \textbf{prior regularization loss}
\begin{equation}
    \mathcal{L}_{\text{prior}} = D_{\mathrm{KL}}\!\bigl(P_{f_s}(c \mid u) \,\|\, P_{f_{s_0}}(c \mid u)\bigr)
\end{equation}
serves as a penalty term that prevents overfitting and preserves the original connective distribution, given the diversity in connective length and frequency.

\paragraph{(2) Curriculum-based optimization.}
To adapt the lightweight model to streaming inference, we employ a curriculum learning strategy. Training proceeds from full-context dialogues to progressively truncated segments, allowing the model to learn to generate coherent turn-transition connectives under partial input. The three coherence-oriented objectives ($\mathcal{L}_{\text{con}},\mathcal{L}_{\text{coh}},\mathcal{L}_{\text{prior}}$) are optimized jointly, ensuring the small model maintains discourse compatibility while gradually adapting to increasingly challenging streaming scenarios, as shown in Eq.~\ref{eq:total}. More details are shown in Appendix~\ref{sec:appendix_implementation}.
\begin{equation}\label{eq:total}
\mathcal{L} = \lambda_{con} \mathcal{L}_{\text{con}} + \lambda_{coh} \mathcal{L}_{\text{coh}} + \lambda_{prior} \mathcal{L}_{\text{prior}}
\end{equation}

\paragraph{(3) Confidence-based connective selection for pre-filling control.}
At inference time, connective emission decisions are usually made under an incomplete ASR context.
To regulate early output, we adopt a Confidence-based connective selection mechanism that selectively emits turn-transition connectives for pre-filling when they are sufficiently reliable.

Given a streaming step $k$, we obtain the top $m$ candidate connective list $C^k=\{c^k_1,... ,c^k_m\}$ generated by the lightweight model. For each connective candidate $c^k = (t^k_1, t^k_2,\dots,t^k_n)$ consisting of $n$ tokens, we estimate token-level uncertainty using predictive entropy.
Specifically, for each token $t_i$ in the connective, we compute its entropy over the vocabulary distribution:
\begin{equation}
H(t^k_i) = - \sum_{v \in \mathcal{V}}
P_{f_s}(v \mid t^k_{<i}, \hat{\mathbf{u}}^{(k)})
\log P_{f_s}(v \mid t^k_{<i},\hat{\mathbf{u}}^{(k)})
\end{equation}

where $\mathcal{V}$ denotes the vocabulary and $P_{f_s}(v \mid t^k_{<i},\hat{\mathbf{u}}^{(k)})$ is the probability of token $v$ given the preceding context under model $f_s$.
We then compute the average entropy ($\bar{H}(c^k)$) over the connective tokens (excluding the special marker token). Finally, the confidence score in streaming step $k$ is obtained by normalizing the entropy with respect to a maximum entropy threshold $H_{max}$:
\begin{equation} 
\mathrm{Conf}(k) = 1 - \frac{\sum_i^m(\bar{H}(c_i^k))}{mH_{\max}} \end{equation}

The system outputs the connective only when $Conf(k)$ exceeds a predefined threshold $\tau$ ($sig^{(k)}=1$). More details are shown in Appendix~\ref{sec:appendix_implementation}.

\section{Experiments}

\subsection{Dataset}

We evaluate the effectiveness of our DDTSR on two widely used benchmarks for daily spoken dialogue: SD-Eval~\citep{ao2024sd} and SpokenNativQA~\citep{alam2025spokennativqa}, both of which are constructed from everyday conversational speech. For SD-Eval, we select samples with unique scripts from three subsets, namely accent (test-acc), age (test-age), and environment (test-env), while excluding the test-emo subset, as it is not well aligned with daily dialogue scenarios. For SpokenNativQA, we adopt the English (en) subset of the official test set. Detailed statistics of the adopted datasets are summarized in Table~\ref{tab:dataset_stats}. Based on the above data, we further divided the dataset into a 8:1:1 training set, validation set, and test set.

\begin{table}[htbp]
\centering
\begin{tabular}{llr}
\toprule
\textbf{Dataset} & \textbf{Subset} & \textbf{\# Samples} \\
\midrule
\multirow{4}{*}{SD-Eval} 
    & test-acc & 445 \\
    & test-age & 507 \\
    & test-env & 284 \\
\cmidrule(lr){2-3}
    & \textit{Total} & \textit{1,236} \\
\midrule
SpokenNativQA & en & 2,322 \\
\midrule
\textbf{Total} & & \textbf{3,558} \\
\bottomrule
\end{tabular}
\caption{Dataset statistics of SD-Eval and SpokenNativQA.}
\label{tab:dataset_stats}
\end{table}

\subsection{Baselines}

\paragraph{End-to-end speech models.}
We select three representative commercial speech models with their APIs: Doubao Realtime, GLM-Realtime, and Qwen3-Omni-Flash-Realtime. 

\paragraph{Cascaded baselines.}
To provide a more controlled comparison, we construct two cascaded baselines with increasing modeling capability. (1) Standard Single-Stream Cascade (SSC). This is a streaming-based FunAudioLLM~\citep{an2024funaudiollm} we modified.
(2) Standard Dual-Stream Cascade (SDC).
To further strengthen the SSC, we construct a dual-stream cascade architecture in which a small connective model and a large language model operate in parallel to generate responses. Unlike our proposed method, this baseline does not incorporate cross-modal interleaving, temporal reordering, or early speech emission.

\subsection{Experimental Implementation}\label{sec:implementation_details}

Our DDTSR system adopts a small–large model architecture. The connective model is initialized from Qwen3-0.6B and fine-tuned to generate fast, non-knowledge-bearing connective utterances, enabling early speech emission with minimal latency. For main content reasoning, we employ two larger backbones, Qwen3-8B and Qwen3-32B, which are accessed via remote APIs to decouple heavy inference from the real-time pipeline. For speech processing, we use an online ASR module based on sherpa-onnx~\footnote{https://github.com/k2-fsa/sherpa-onnx}, and adopt CosyVoice2~\citep{du2024cosyvoice} for incremental TTS synthesis. The connective model, streaming ASR, and streaming TTS are all deployed locally on a single RTX 4090D GPU.

\begin{table*}[!t]
\centering
\resizebox{\textwidth}{!}{%
\begin{tabular}{ccccccccccccccc} 
\toprule
\multirow{3}{*}{\textbf{Dataset}} 
& \multirow{3}{*}{\textbf{Strategy}} 
& \multirow{3}{*}{\textbf{Large Model}}
& \multicolumn{9}{c}{\textbf{Latency (ms)}} \\
\cmidrule(lr){4-12}
&  
&  
& \multicolumn{3}{c}{\textbf{Perception}} 
& \multicolumn{3}{c}{\textbf{Reaction}} 
& \multicolumn{3}{c}{\textbf{Waiting}} \\
\cmidrule(lr){4-6} \cmidrule(lr){7-9} \cmidrule(lr){10-12}
&  
&  
& \textbf{Opt.} & \textbf{Rem.} & \textbf{Avg.}
& \textbf{Opt.} & \textbf{Rem.} & \textbf{Avg.}
& \textbf{Opt.} & \textbf{Rem.} & \textbf{Avg.} \\
\midrule
\multirow{8}{*}{SD-Eval}
& SSC & \multirow{3}{*}{8B}
& - & - & 388
& - & - & 615
& - & - & 1003 \\
& SDC &
& 363 & 438 & 391
& 416 & 559 & 470
& 779 & 997 & 861 \\
& DDTSR &
& 90 & 327 & \textbf{113}
& 428 & 528 & \textbf{435}
& 515 & 855 & \textbf{548} \\
\cmidrule(lr){2-12}
& SSC & \multirow{3}{*}{32B}
& - & - & 388
& - & - & 736
& - & - & 1124 \\
& SDC &
& 354 & 443 & 387
& 422 & 725 & 536
& 776 & 1168 & 923 \\
& DDTSR &
& 89 & 276 & \textbf{102}
& 430 & 677 & \textbf{447}
& 519 & 953 & \textbf{549} \\
\cmidrule(lr){2-12}
& GLM-realtime & -
& - & - & -
& - & - & -
& - & - & 763 \\
& Qwen3-omni-flash-realtime & -
& - & - & -
& - & - & -
& - & - & 626 \\
& Doubao-realtime & -
& - & - & -
& - & - & -
& - & - & \textbf{352} \\
\midrule
\multirow{8}{*}{SpokenNativQA}
& SSC & \multirow{3}{*}{8B}
& - & - & 308
& - & - & 624
& - & - & 931 \\
& SDC&
& 261 & 312 & 311
& 443 & 546 & 544
& 704 & 858 & 854 \\
& DDTSR &
& 111 & 309 & \textbf{235}
& 417 & 559 & \textbf{506}
& 528 & 869 & \textbf{741} \\
\cmidrule(lr){2-12}
& SSC & \multirow{3}{*}{32B}
& - & - & 307
& - & - & 714
& - & - & 1022 \\
&SDC&
& 312 & 308 & 308
& 448 & 689 & 683
& 760 & 997 & 991 \\
& DDTSR &
& 120 & 305 & \textbf{236}
& 436 & 686 & \textbf{592}
& 555 & 991 & \textbf{828} \\
\cmidrule(lr){2-12}
& GLM-realtime & -
& - & - & -
& - & - & -
& - & - & 763 \\
& Qwen3-omni-flash-realtime & -
& - & - & -
& - & - & -
& - & - & 634 \\
& Doubao-realtime & -
& - & - & -
& - & - & -
& - & - & \textbf{366} \\
\midrule
\end{tabular}
}
\caption{
External latency breakdown across datasets and strategies (reported in milliseconds).
SSC, SDC, and DDTSR are evaluated under both 8B and 32B model scales,
while GLM-realtime and Doubao-realtime are model-agnostic. Opt. and Rem. report the average audio latency with and without connective generation, respectively.
}
\label{tab:latency_model_dataset}
\end{table*}

\subsection{Evaluation Metrics}

We evaluate DDTSR from two complementary dimensions: latency-oriented metrics and quality-oriented metrics. The metrics are specifically designed for streaming low-latency spoken interaction, aiming to quantify both real-time responsiveness and output coherence. 

\textbf{Latency-Oriented Metrics.} To precisely characterize where latency arises in streaming interaction, we decompose system delay into three stages: (1) Perception Latency: time from the completion of sending the last user audio chunk to the model begins responding with normal (content-carrying) output. This metric isolates the perception stage and removes the influence of input utterance length. (2) Reaction Latency: time from the model begins responding with normal (content-carrying) output to the emission of the first synthesized audio chunk. It combines the cost of LLM reasoning and initial TTS synthesis, which is commonly reported as response latency in prior work.
(3) Waiting Latency: The sum of the perception latency and reaction latency. This directly reflects the perceived delay experienced by users.

\textbf{Quality-Oriented Metrics.}
Since all compared methods share the same large language model backbone, answer correctness is controlled by design. We therefore focus on whether early connectives affect interaction quality: (1) Textual Coherence: assessed using an LLM-as-a-judge protocol (G-Eval~\citep{liu2023g}) to measure logical consistency and discourse coherence~\citep{li2021addressing, stede2012discourse}.
(2) Speech Naturalness: evaluated with UTMOSv2~\citep{baba2024utmosv2}, an objective metric highly correlated with human perception.

The evaluation details are shown in the Appendix~\ref{app: Evaluation Metrics}.

\section{Results and Analysis}

\paragraph{Main Results in Latency Reduction.}
Table~\ref{tab:latency_model_dataset} reports the latency breakdown of different strategies across datasets and model scales. Overall, DDTSR consistently achieves the lowest latency in all settings, reducing waiting latency by a large margin compared with standard cascaded baselines (SSC and SDC). On SD-Eval with the 8B model, DDTSR reduces average waiting latency from 1003 ms (SSC) and 861 ms (SDC) to 548 ms, corresponding to a 45.4\% reduction. On the Opt. setting with connector phrase generation, waiting latency drops to as low as 515 ms. Similar gains are observed with the 32B model. On SpokenNativQA, DDTSR also achieves substantial improvements (17.5\%), reducing waiting latency from 931 ms to 741 ms (8B) and from 1022 ms to 828 ms (32B). These results demonstrate that DDTSR effectively brings cascaded systems into the sub-second response regime, approaching the responsiveness of end-to-end real-time speech models.

\paragraph{Reduction in Reaction Latency.}
A key contributor to the overall improvement is the substantial reduction in reaction latency enabled by the connective-guided small-large model synergy. Compared with the standard cascaded baseline (SSC), both SDC and DDTSR achieve notable reductions in reaction time. For example, on SD-Eval with the 8B model, reaction latency decreases from 615 ms (SSC) to 470 ms with SDC and further to 435 ms with DDTSR. By replacing the long “think-then-speak” bottleneck with parallel small–large model execution, the system significantly shortens the reaction time to first response.

\begin{figure*}[t]
    \centering
    \includegraphics[width=\linewidth]{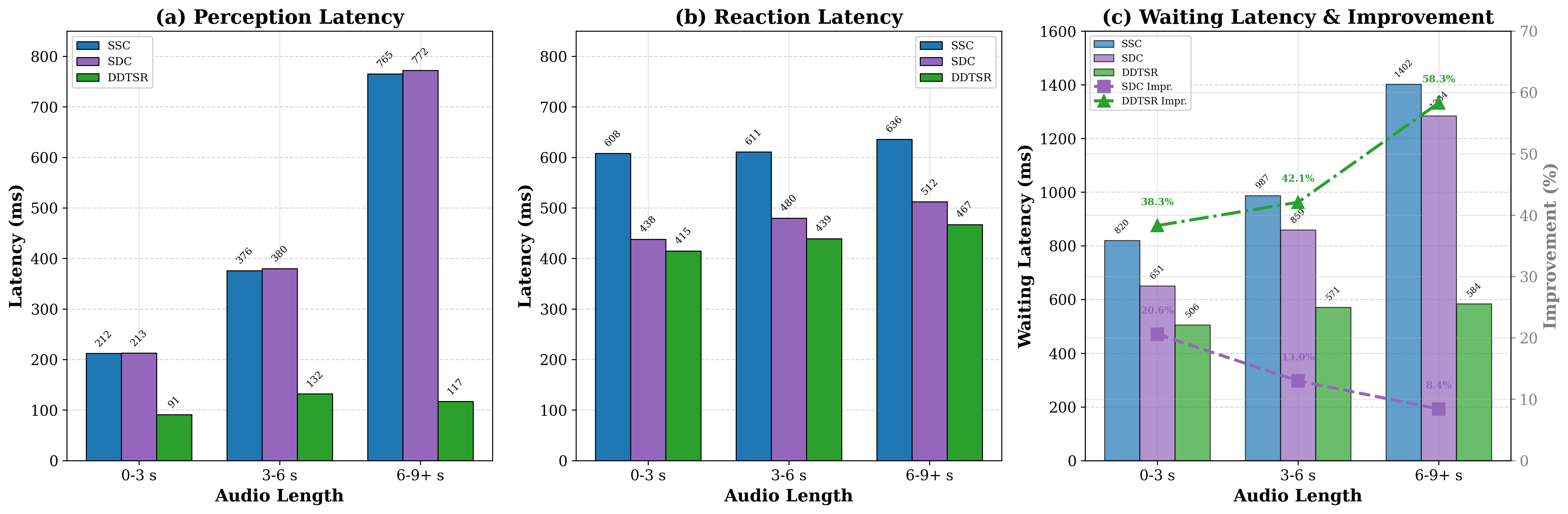}
    \caption{Latency analysis on SD-Eval with respect to input audio length.}
    \label{fig:audio_length_latency}
\end{figure*}

\paragraph{Reduction in Perception Latency.}
Through the proposed cross-modal collaboration mechanism, DDTSR further reduces perception latency. As shown in Table~\ref{tab:latency_model_dataset}, perception latency on SD-Eval drops from approximately 388 ms under SSC and SDC to only 113 ms with the 8B model and 102 ms with the 32B model. On the Opt. setting with connective generation, perception latency further reaches as low as 89 ms. This improvement is achieved because the streaming architecture allows the small model to begin reasoning while the user is still speaking, rather than waiting for ASR finalization. By overlapping partial ASR processing with early connective generation, the system substantially shortens the time spent in the perception stage, leading to a much faster transition from listening to speaking.

\paragraph{Effect of Model Scale and Datasets.}
The latency advantages of DDTSR remain robust across different model scales and datasets. Owing to the proposed small–large model synergy, the system is able to maintain consistently low latency (500 ms in SD-Eval and 800 ms in SpokenNativQA) regardless of the capacity of the underlying large model. This is because early response emission primarily relies on the lightweight model, whose inference time is largely insensitive to the scale of the main reasoning model. 

In addition, latency variations across datasets are mainly attributed to differences in the frequency and distribution of connective connectives. Datasets with more frequent (94\% in SD-Eval and 38\% in SpokenNativQA) turn-initial connectives naturally allow earlier commitments and thus achieve lower effective latency, whereas datasets with fewer such opportunities exhibit relatively higher waiting times. Importantly, the likelihood of generating such connectives is also scenario-dependent, making this strategy particularly well-suited to casual spoken interactions such as SD-Eval.

\begin{table}[!t]
\centering
\resizebox{\linewidth}{!}{%
\begin{tabular}{ccccc}
\toprule
\multirow{2}{*}{\textbf{Dataset}} 
& \multirow{2}{*}{\textbf{Strategy}} 
& \multirow{2}{*}{\textbf{Model}} 
& \textbf{Text Level} 
& \textbf{Speech Level} \\
\cmidrule(lr){4-4} \cmidrule(lr){5-5}
 &  &  & \textbf{Cons. / Cohe.} & \textbf{UTMOSv2} \\
\midrule
\multirow{7}{*}{SD-Eval} 
& Gold Standard & -- & 4.77 / 4.43 & -- \\
\cmidrule(lr){2-5}
& SSC  & \multirow{3}{*}{8B}  & 4.76 / 4.54 & 3.16 \\
& SDC  &                     & 4.77 / 4.46 & 3.20 \\
& DDTSR&                     & 4.77 / 4.49 & 3.12 \\
\cmidrule(lr){2-5}
& SSC  & \multirow{3}{*}{32B} & 4.94 / 4.71 & 3.09 \\
& SDC  &                     & 4.78 / 4.59 & 3.07 \\
& DDTSR&                     & 4.87 / 4.57 & 3.10 \\
\midrule
\multirow{7}{*}{SpokenNativQA} 
& Gold Standard & -- & 4.70 / 4.39 & -- \\
\cmidrule(lr){2-5}
& SSC  & \multirow{3}{*}{8B}  & 4.72 / 4.54 & 3.14 \\
& SDC  &                     & 4.71 / 4.50 & 3.13 \\
& DDTSR&                     & 4.63 / 4.43 & 3.14 \\
\cmidrule(lr){2-5}
& SSC  & \multirow{3}{*}{32B} & 4.69 / 4.56 & 3.06 \\
& SDC  &                     & 4.72 / 4.56 & 3.08 \\
& DDTSR&                     & 4.61 / 4.43 & 3.08 \\
\bottomrule
\end{tabular}
}
\caption{
Text- and speech-level quality comparison across different strategies on SD-Eval and SpokenNativQA. Text quality is reported as Consistency/Coherence (“Cons./Cohe.”) via an LLM-as-a-judge protocol, and speech quality by UTMOSv2. Gold Standard refers to the ground-truth QA pairs.
}
\label{tab:response_quality_combined}
\end{table}

\paragraph{Output Quality of DDTSR.}
Beyond latency reduction, we further examine whether DDTSR affects output quality. As shown in Table~\ref{tab:response_quality_combined}, DDTSR maintains performance comparable to baseline strategies across both text and speech modalities, and remains competitive relative to the Gold Standard reference. For example, on SD-Eval with the 8B model, DDTSR achieves a consistency score of 4.77 and a coherence score of 4.49, nearly identical to SSC (4.76 / 4.54) and SDC (4.77 / 4.46). In terms of speech quality, UTMOSv2 scores of DDTSR are also on par with baselines (e.g., 3.12 vs. 3.16 on SD-Eval with 8B), indicating that concatenating TTS outputs from two tracks does not degrade perceptual naturalness.

\paragraph{Audio Length Effects on Improvement Ratio.}
To examine how latency improvements vary with input duration, we stratify SD-Eval into three audio-length ranges (0--3\,s / 3--6\,s / 6--9+\,s) and conduct a detailed statistical analysis, as shown in Figure~\ref{fig:audio_length_latency}. The results reveal three key findings. 
(1) The perception latency of DDTSR remains largely stable (around 100 ms) regardless of input length, whereas baseline systems exhibit a clear increase as utterances become longer. This robustness is attributed to the cross-modal collaboration mechanism, which allows the small model to initiate processing before ASR finalization, as illustrated in Figure~\ref{fig:audio_length_latency}(a). 
(2) Although reaction latency shows relatively modest growth for all systems, DDTSR consistently achieves larger reductions compared with SSC and SDC. This additional gain stems from the small–large model synergy, which effectively shortens reasoning time, as reflected in Figure~\ref{fig:audio_length_latency}(b). 
(3) Overall, the waiting latency reduction brought by DDTSR becomes more pronounced for longer inputs (from 38.3\% to 58.3\%), indicating that the proposed framework yields increasing benefits in extended interactions. This property is particularly valuable for long-form and multi-turn dialogue scenarios.

\section{Conclusion}
We present DDTSR, a discourse-aware dual-track streaming framework for low-latency spoken dialogue. Through connective-guided small–large model synergy and streaming-based cross-modal collaboration, the system can initiate safe and natural responses at the earliest speakable moment while allowing reasoning to continue in parallel. A curriculum-learning-based coherence enhancement mechanism further ensures that early emissions remain consistent with subsequent responses. Experiments demonstrate that DDTSR consistently reduces perceived latency by 19\%--51\%, without requiring modifications to existing ASR or TTS modules. Meanwhile, response quality in both text and speech modalities is effectively preserved. We believe DDTSR provides a practical and general solution toward truly interactive speech agents, and opens new directions for discourse-aware and latency-sensitive dialogue system design.





\clearpage
\bibliographystyle{named}
\bibliography{ijcai26}

\clearpage
\appendix

\section{Appendix}
\label{sec:appendix}

\subsection{Perplexity Computation}
\label{app:ppl}

Given a response sequence $R = (r_1, \dots, r_{|R|})$ with $|R_L|$ tokens, the perplexity under model $f_s$ conditioned on context $u$ is defined as:
\begin{equation}
    \mathrm{PPL}_{f_s}(c, R_L \mid u) = \exp\!\left(-\frac{1}{|R_L|} \sum_{t=1}^{|R_L|} \log P_{f_s}(r_t \mid u, c, r_{<t})\right),
\end{equation}
where $r_{<t} = (r_1, \dots, r_{t-1})$ denotes the preceding tokens.

\subsection{Connective Data Construction}\label{app:connective_data_construction} 

\subsubsection{Connective Definition} 
To support low-risk early speech and enable rapid response onset, we characterize the connectives used in this work by two essential properties. First, \textbf{positional specificity}. A connective must appear turn-initially at the boundary between two consecutive speakers, explicitly linking Speaker~B’s response ($S_2$) to the immediately preceding utterance produced by Speaker~A ($S_1$). We therefore focus on turn-transition connectives that form an $S_1$–[connective]–$S_2$ structure, excluding sentence-internal or mid-utterance discourse markers.
Second, \textbf{non--truth-conditional meaning}. The connective must not contribute to the propositional content of $S_2$. Removing the connective preserves the semantic payload of the response and only affects how the discourse relation between the two turns is interpreted. This property makes such connectives suitable as low-risk prefixes that can be emitted before substantive reasoning is complete. Based on these two properties, we ground our definition in Fraser’s theory of pragmatic markers~\citep{fraser2009account} and selectively retain discourse markers that meet these constraints. Concrete categories and examples are provided in the Table ~\ref{tab:fraser_connective_types}.

\begin{table*}[t]
\centering
\small
\renewcommand{\arraystretch}{1.3}
\begin{tabular}{>{\raggedright\arraybackslash}p{2.5cm} 
                >{\raggedright\arraybackslash}p{2.8cm} 
                >{\raggedright\arraybackslash}p{3.2cm} 
                >{\raggedright\arraybackslash}p{7cm}}
\toprule
\textbf{Fraser's Category} & \textbf{Subtype} & \textbf{Connective Examples} & \textbf{Function in Turn Transition} \\
\midrule
\multirow{3}{=}{\textbf{Discourse Markers (DMs)}} 
& Contrastive (CDM) 
& \textbf{but, however, nevertheless}
& Speaker B introduces a contrast, objection, or exception to A's utterance. \\[0.5ex]
& Elaborative (EDM) 
& \textbf{and, moreover, besides}
& Speaker B adds information or provides elaboration related to A's utterance. \\[0.5ex]
& Inferential (IDM) 
& \textbf{so, therefore, thus}
& Speaker B derives a conclusion or consequence based on A's utterance. \\
\midrule
\multirow{3}{=}{\textbf{Discourse Structure Markers (DSMs)}} 
& Topic-Orienting (TOM) 
& \textbf{by the way, speaking of which}
& Speaker B introduces a related subtopic, managing topic transitions. \\[0.5ex]
& Discourse-Managing (DMM) 
& \textbf{first, in summary, in other words}
& Speaker B comments on or organizes the structure of upcoming discourse. \\[0.5ex]
& Attention-Managing (AM) 
& \textbf{look, listen, hey}
& Speaker B directs or secures the attention of A before proceeding. \\
\bottomrule
\end{tabular}
\caption{Turn-transition connective types adapted from Fraser's discourse marker classification system. Connectives occur at the beginning of Speaker B's turn and do not contribute to truth conditions.}
\label{tab:fraser_connective_types}
\end{table*}

\subsubsection{Connective Construction} 
To train the lightweight model for connective-aware generation, we construct a task-specific connective-annotated dialogue dataset targeting turn-transition connectives defined in Section~3.1. The dataset is built through a two-stage automatic pipeline that balances precision and coverage. First, we extract naturally occurring turn-initial, non-truth-conditional connectives using POS-based heuristics, producing high-precision annotations. Second, for turns without explicit connectives, an LLM is prompted with our connective definition to determine whether a connective is warranted and to generate an appropriate one if necessary. The resulting dataset pairs each dialogue context with a connective and its corresponding continuation, enabling the lightweight model to learn early connective generation that remains compatible with subsequent large-model responses.

\paragraph{POS-Based Connective Extraction.}  
To construct an initial set of turn-transition connectives from naturally occurring dialogue, we employ a POS-based extraction strategy. Each response is first segmented using punctuation, and candidate prefixes are tested incrementally. A prefix is recognized as a connective if all its tokens satisfy one of the following patterns: Type A markers, consisting of interjections (INTJ), adverbs (ADV), pronouns (PRON), auxiliary verbs (AUX), meta-verbs (VERB), or particles (PART); Type B markers, consisting of adjectives (ADJ), adverbs (ADV), abstract nouns (NOUN), or subordinating conjunctions (SCONJ); or a mixed Type A+B pattern combining features of both. The prefix must also contain no substantial content such as proper nouns, concrete entities, or content verbs, and must meet minimal length requirements. Type B markers are further constrained to include at least one adjective to capture evaluative or reactive expressions. This extraction produces a high-precision set of linguistically valid connectives for model training and evaluation, as illustrated in Figure~\ref{alg:pos_extraction}.

\begin{figure}[htbp]
\centering
\begin{minipage}{0.9\linewidth} 
\begin{algorithm}[H]
\caption{POS-Based connective extraction pipeline.}
\label{alg:pos_extraction}
\textbf{Input:} Text $T$ \\
\textbf{Output:} Connective $C$, Remainder $R$
\begin{algorithmic}[1]
\STATE Initialize $C \gets \emptyset$, $R \gets T$
\STATE $segments \gets \text{split\_by\_punctuation}(T)$

\FOR{each segment $s$ in $segments$}
    \STATE $candidate \gets C + s$
    \STATE $tokens \gets \text{dependency\_parse}(candidate)$
    \IF{$\text{has\_substantial\_content}(tokens)$}
        \STATE \textbf{break} \COMMENT{Stop accumulation}
    \ELSIF{all tokens $\in$ allowed POS set}
        \STATE $C \gets candidate$ \COMMENT{Accumulate connective}
    \ELSE
        \STATE \textbf{break}
    \ENDIF
\ENDFOR

\STATE $R \gets T \setminus C$ \COMMENT{Remove connective from original text}
\STATE \textbf{return} $(C, R)$
\end{algorithmic}
\end{algorithm}
\end{minipage}
\end{figure}

\paragraph{LLM-Prompted Connective Generation.}  
To generate connective-annotated samples, we design an LLM prompt that operationalizes the definition of turn-initial, non–truth-conditional connectives. The prompt instructs the model to analyze the preceding and current utterances (S1 and S2), determine whether a connective is present, and, if absent, produce an appropriate connective from a predefined functional category space (CDM, EDM, IDM, TOM, DMM, AM, NONE). The output is strictly structured, facilitating automatic annotation while ensuring consistency with the discourse-level characteristics required for early, low-risk connective pre-filling. Calibration examples are optionally included to illustrate typical and boundary cases. The format and details of the prompt are shown in Figure ~\ref{fig:llm_prompt}.

\begin{figure*}[htbp]
\centering
\begin{tcolorbox}[colback=gray!10, colframe=gray!50, title=LLM Connective Prompt Format, sharp corners, width=\textwidth]
\small\ttfamily
\textbf{Task Framing and Input:} \\
You are a discourse and conversation analysis expert.\\[1mm]
\textbf{Input:} \\
S1 (Speaker A): <S1> \\
S2 (Speaker B): <S2> \\[1mm]

\textbf{Target Phenomenon Definition:} \\
Identify turn-taking discourse connectives at the beginning of S2. \\
A connective is turn-initial, non-truth-conditional, span-based, \\
and introduces no new propositional content. \\[1mm]

\textbf{Functional Category Space:} \\
CDM | EDM | IDM | TOM | DMM | AM | NONE \\[1mm]

\textbf{Decision Procedure:} \\
1. Check whether S2 begins with a connective. \\
2. If not, generate an appropriate connective or NONE. \\[1mm]

\textbf{Output Specification and Calibration:} \\
Output exactly two lines: \\
CONNECTIVE PRESENT: YES | NO \\
CONNECTIVE: <CONNECTIVE or NONE> \\[1mm]

Optional calibration examples illustrating typical and boundary cases.
\end{tcolorbox}
\caption{LLM prompt format for turn-transition connective generation.}
\label{fig:llm_prompt}
\end{figure*}

\subsubsection{Connective Dataset Statistics} \label{app:connective_stats} To further characterize the constructed connective datasets, we report several descriptive statistics in Table~\ref{tab:connective_stats}. 
These statistics provide insights into both dataset scale and label distribution, which are critical for assessing the coverage and diversity of discourse connectives.

\begin{itemize}
    \item \textbf{Total Samples}: the total number of samples in the dataset, indicating overall dataset size.
    \item \textbf{Samples with Connectives}: the number of samples in which a connective occurs, reflecting how many instances contribute to connective learning.
    \item \textbf{Connective Types}: the number of unique connective labels, showing the label space richness.
    \item \textbf{Normalized Label Entropy}: computed as the label entropy divided by the logarithm of the number of connective types. It quantifies how uniformly the different connectives are distributed across the dataset; higher values indicate more balanced and less skewed usage.
\end{itemize}
Table~\ref{tab:connective_stats} summarizes these statistics. Together, they show that our connective datasets are not only sufficiently large, but also exhibit a well-utilized and relatively balanced connective label space, supporting stable training and generalization. 

\begin{table}[h]
\centering
\small
\begin{tabular}{lcccc}
\toprule
\textbf{Dataset} & \textbf{\#Samp.} & \textbf{\#Conn.} & \textbf{\#Types} & \textbf{Entropy} \\
\midrule
SD-Eval      & 1236 & 1125 & 240 & 0.75 \\
SpokenNativQA & 2322 & 1055 & 146 & 0.65 \\
\bottomrule
\end{tabular}
\caption{Statistics of the constructed connective datasets. \#Samp.: total number of samples; \#Conn.: number of samples containing connectives; \#Types: number of distinct connective types; Entropy: the normalized entropy of connective label distribution (higher value indicates more uniform label usage).}
\label{tab:connective_stats}
\end{table}

\subsection{Experimental Implementation}\label{sec:appendix_implementation}
We adopt a small--large model configuration for DDTSR. Specifically, the connective model is built upon \textbf{Qwen3-0.6B} and further \textbf{fine-tuned} for fast non-knowledge-bearing connective connective generation. 
For the main content reasoning and response generation, we evaluate two larger backbones, \textbf{Qwen3-8B} and \textbf{Qwen3-32B}, which are accessed via API calls to decouple complex inference from the on-device real-time pipeline.

For streaming ASR, we implement an online recognition module using \texttt{sherpa-onnx} with a streaming Zipformer backend. 
For streaming TTS, we employ \textbf{CosyVoice2} to synthesize and play back speech incrementally.
In our deployment, the \textbf{connective model}, \textbf{streaming ASR}, and \textbf{streaming TTS} are all hosted locally on a single \textbf{NVIDIA RTX 4090D}, while the large model is served remotely via API.

\begin{table}[htbp]
\centering
\resizebox{\linewidth}{!}{%
\begin{tabular}{llll}
\toprule
\textbf{Component} & \textbf{Role} & \textbf{Model} & \textbf{Deployment} \\
\midrule
\multirow{3}{*}{LLM} 
& Connective (connective) 
& Qwen3-0.6B (fine-tuned) 
& Local \\
& \multirow{2}{*}{Main content (reasoning)} 
& Qwen3-8B 
& \multirow{2}{*}{Remote} \\
& 
& Qwen3-32B 
& \\
\midrule
Streaming ASR 
& Incremental transcription 
& sherpa-onnx-streaming-zipformer-en-2023-06-26 
& Local \\
\midrule
Streaming TTS 
& Incremental synthesis 
& CosyVoice2 
& Local \\
\bottomrule
\end{tabular}%
}
\caption{System configuration and deployment setup.}
\label{tab:system_config}
\end{table}

\paragraph{Curriculum-Learning-Based Discourse Continuity Enhancement.} 
We implement this module through three training stages. \textbf{Parameter-efficient adaptation} fine-tunes the lightweight model with LoRA ($r{=}32$, $\alpha{=}64$; learning rate $2{\times}10^{-5}$) to jointly generate turn-transition connectives and full responses. \textbf{Curriculum scheduling} organizes samples into four stages based on the empirical distribution of audio chunks (500 ms each), reflecting response difficulty under partial input; training proceeds from harder to easier conditions, with epochs allocated in reverse order (5/3/3/2). The weighting parameters $\lambda_{\text{con}}$, $\lambda_{\text{coh}}$, and $\lambda_{\text{prior}}$ are set to 1.0, 0.5, and 0.1, respectively. \textbf{Confidence-based pre-filling} bridges training and inference by emitting early connectives whose confidence exceeds dataset-specific thresholds ($\tau{=}0.45$ for SD-Eval, $0.15$ for SpokenNativQA), with the pre-filling horizon capped at $H_{\max}{=}2$. We select the top 5-10 candidate connectives as $C^k$ in the step $k$ and found that 5 achieved the best performance.

\subsection{Evaluation Metrics}
\label{app: Evaluation Metrics}
For text quality, we focus on logical consistency and discourse coherence. Logical Consistency measures whether the generated response aligns with the dialogue context without contradictions~\citep{li2021addressing, liualigning}. discourse coherence evaluates discourse-level organization and fluency across sentences~\citep{grosz1986attention, stede2012discourse, fabbri2021summeval}. Following recent work on automatic evaluation, we adopt an LLM-as-a-judge protocol using the G-Eval framework~\citep{liu2023g}, which has been shown effective for assessing discourse quality.

\end{document}